# Multi-stream Data Analytics for Enhanced Performance Prediction in Fantasy Football

Nicholas Bonello, Joeran Beel, Seamus Lawless, Jeremy Debattista

School of Computer Science and Statistics, Trinity College Dublin, Ireland
`bonellon@tcd.ie, beelj@scss.tcd.ie, seamus.lawless@scss.tcd.ie,`
`jeremy.debattista@adaptcentre.ie,`

**Abstract.** Fantasy Premier League (FPL) performance predictors tend to base their algorithms purely on historical statistical data. The main problems with this approach is that external factors such as injuries, managerial decisions and other tournament match statistics can never be factored into the final predictions. In this paper, we present a new method for predicting future player performances by automatically incorporating human feedback into our model. Through statistical data analysis such as previous performances, upcoming fixture difficulty ratings, betting market analysis, opinions of the general-public and experts alike via social media and web articles, we can improve our understanding of who is likely to perform well in upcoming matches. When tested on the English Premier League 2018/19 season, the model outperformed regular statistical predictors by over 300 points, an average of 11 points per week, ranking within the top 0.5% of players - rank 30,000 out of over 6.5 million players.

## 1  Introduction

Fantasy premier League (FPL) [1] is the official fantasy football browser game for the English premier league, with over 6.5 million players actively competing against each other every season aiming to accumulate the highest number of points. Fantasy sports are continuously becoming more popular, with Fantasy Premier League leading the way in terms of number of players per season. Because of this rise in popularity, achieving a respectable rank is becoming increasingly difficult, much less trying to win on either the official or paid leagues.

In this paper we present a novel approach for recommending FPL player selections for upcoming gameweeks that automatically incorporates human feedback into the model through publicly available online data such as blogs written by domain experts (Fig. 1. By considering news articles, fantasy-football prediction blogs and even tweets using fantasy premier league specific hashtags, our recommender system is able to consider feedback from experts and fans alike. These are used as additional parameters in our predictive model that enables our model to automatically incorporate external factors such as midweek game lineups and performances, managerial decisions, rotations and injuries together with unexpectedly good or bad performances in previous weeks. We found that incorporating different data sources significantly improved our various models final rank. The statistical model placed within the top 30% of players, improving to top 0.5% overall rank (within top 30,000 out of 6.5 million players) when incorporating data from betting markets and blog posts. Our recommender system is the first of its kind to quantify the added value of incorporating public textual data into our predictions.

[1] https://fantasy.premierleague.com/

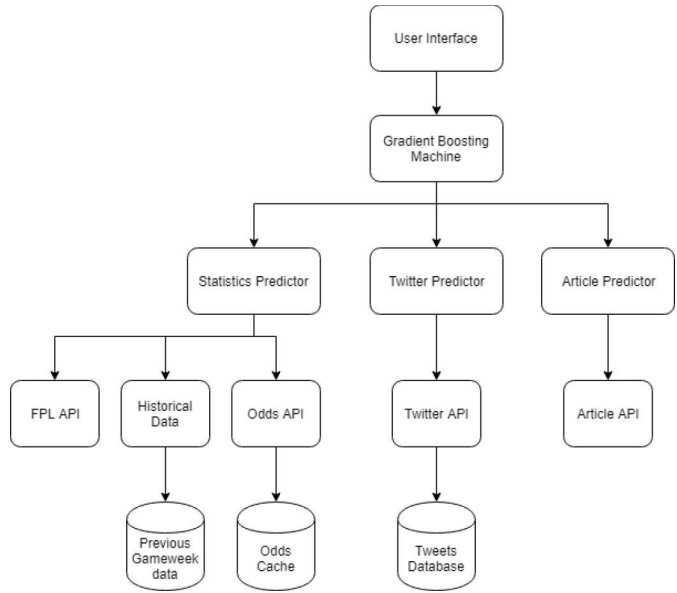

**Fig. 1.** Predictive Model flow diagram

## 2 Related Work

The general inclination observed in previous works has been to use purely historical statistical data in combination with Naive-Bayes, SVMs, Random Forests and other ensemble methods to predict future player performances. Using strictly historical statistical data to train a Gaussian Naive Bayes algorithm, Thapliya was able to predict future player performances with a reported accuracy of 81% [6]. Using similar training data, Raghunandh found that using ensemble methods such as Gradient Boosting Machines (GBM) could help, reporting an average precision of 86% across the duration of an entire season when predicting player performances [5]. Bonomo et. Al derived a mathematical model attempting to predict ideal line-ups per gameweek in the Argentinian first division [1]. Very similar to our goal of predicting the optimal weekly line-ups by incorporating previous information with knowledge gained from press-reports and manager updates before games. The aim of this model is to build an index that can produce reasonable representations of what will happen in future gameweeks. In more sophisticated models, Matthews presented an innovative fantasy football predictor that consisted of belief-state MDP models combined with Bayesian Q-Learning algorithms to train models on the past five years of football data [3]. Their most successful model used a state of the art Bayesian Q-learning model to handle the uncertainty, placing the machine within the top 500 players out of the 2.5million participants (0.01%). The original model was naive; acting myopically - only considering the single next gameweek. Even with all these restrictions, this model still achieved a very respectable rank of 113,921. By extending this model to also look back on data from the previous season, the model leaves out all the players who did not appear but still improved performance slightly, upping the rank to 60,633.

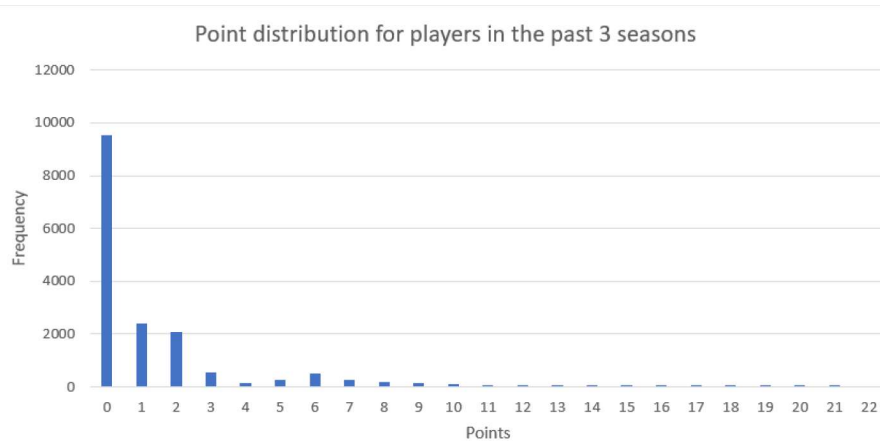

**Fig. 2.** Points Frequency Distribution for players in past 3 years of FPL data. The large majority of players get 0-2 points per game, meaning that training a model tasked with predicting players who will score a large number of points consistently requires an algorithm suitable for such unbalanced datasets.

## 3 Methods

### 3.1 Gradient Boosting Machines

Raghunandh reported that while the different tree models all proved to be effective, gradient-boosting machines (GBM) proved to be the most optimal for fantasy sport predictions [5]. We report similar results, with ensemble methods showing far higher suitability to the task when compared with alternate supervised learning approaches such as SVMs. We found that GBMs are the most effective in this case, particularly due to the unbalanced data we deal with (Fig. 2).

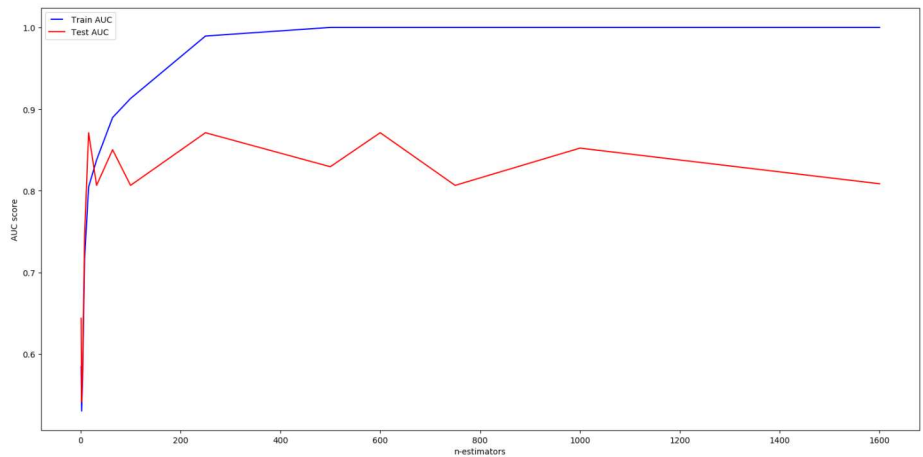

**Fig. 3.** The Area Under Curve - Receiver Operating Characteristics (AUC-ROC) curve is plotted for various parameter values. This allows us to visually identify how well the model can distinguish classes by comparing the different AUC-ROC scores, identifying which value is optimal while preventing overfitting [2].

### 3.2 Dataset

As shown by the predictive model architectural diagram shown in Fig. 1, the dataset we created is a combination of all the statistical data available from the FPL API [2], betting odds retrieved through football-api [3], tweets scraped using the tweepy library [4] and English blog-post sentiment scores for the top 100 search results per query through the Aylien API [5] for each player, per previous match in the current football season. The dataset is publicly available and updated weekly [6]. The application automatically scrapes all the latest available data from the different data sources whenever a new gameweek starts, removing all injured players and players who did not make an appearance in the previous week. An additional binary valued column - "isCaptain" is added, having a value of '1' for all players having more than 6 points in the current gameweek, and '0' for all other players. This column is the predictor variable that we attempt to predict for upcoming gameweeks. The dataset is then split according to the player positions; goalkeeper, defender, midfielder or forward. Depending on this role, different factors will determine whether a player is in good form. For example, defensive odd statistics are important for defenders while useless for forwards. By visualising the importance of variables from previous model training (Fig.

---

[2] https://fantasy.premierleague.com/
[3] https://rapidapi.com/api-sports/api/api-football
[4] https://www.tweepy.org/
[5] https://aylien.com/text-api/
[6] Anonymous

5), the most important factors can be identified. Once training is complete, all the statistical, betting, and textual data for the upcoming gameweek fixtures are retrieved and passed on to the model for predictions.

### 3.3 Implementation

Multiple different approaches were considered and tested. Most notably different machine learning algorithms including SVMs, Random Forests and GBMs showed the most promise. However, GBMs were finally picked due to their robustness, adaptability and proven success when predicting future FPL scores as recommended by Raghunandh [5]. Models were then tested using different data sources to evaluate how significant combining different data sources can be. Ultimately we found that including tweets was not viable due to standard sentiment analysis methods having low performance accuracy, not being able to properly evaluate the sentiment due to spelling mistakes, emoji usages and less focus on correct grammar (e.g. Fig. 4). Even when tweets were grammatically correct, the football language could not be correctly understood by standard sentiment analysis libraries. On the other hand, blogs and news articles could be more accurately understood, likely because of their length, correct usage of grammar and significantly less frequent spelling mistakes. Through named entity extraction (NEE) we could identify the players being mentioned in these blogs and calculate the authors sentiment on the topic (Entity based sentiment analysis) using the Aylien API [7] [4]. A web scraper with predefined FPL specific queries is run and the top 100 results are analyzed. We found that considering more blogs does not add any further value to the model and even begins to decrease accuracy due to excess noise. This technique will be repeated every week of the season, with the aim of predicting the best possible lineup per gameweek.The models will be tested by running them on all the gameweeks in the Premier League 2018/19 season, with all previous gameweek data made available to each instance.

The upcoming fixtures market odds are then scraped and the odds converted to likelihood probabilities. Player specific markets such as "player to score" were not available but could have helped in identifying the individual probabilities rather than the entire team scoring probability.

It is important to note that the rules followed by the models differed slightly from the real game. Since we present a recommender system, our model selects what it believes to be the optimal lineup every week, but real FPL players are restricted and can only transfer one new player per week without any point penalties. This difference means that our recommender system should be used as an assistant providing accurate recommendations per week.

---

[7] https://aylien.com/text-api/

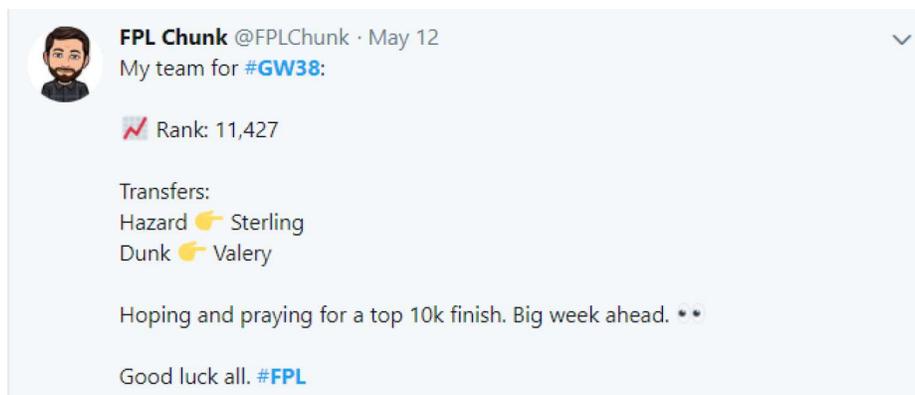

**Fig. 4.** Example Tweet using emojis to represent sentiment. While sentiment is easy for human readers to understand, an algorithm designed to convert tweet text into a sentiment rating requires the algorithm to be able to understand the context behind an emoticon or abbreviation.

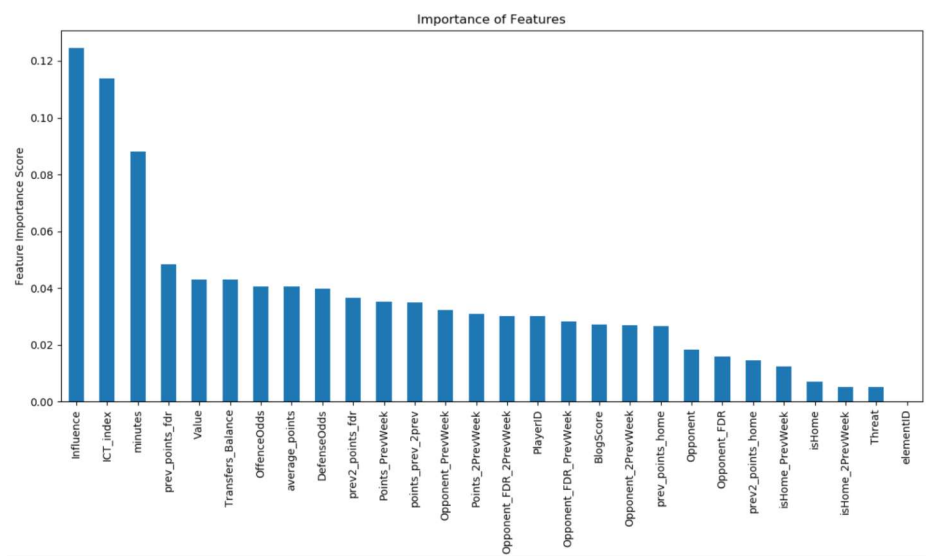

**Fig. 5.** Importance of variables for goalkeeper prediction. The most important variables are influence, minutes and the points gained in the previous week. Interestingly, being home or away seems to have little to no importance to the model.

## 4 Results

The final GBM was evaluated by comparing the number of points it could achieve in individual gameweeks to purely statistical models - representing the baseline. As shown in Fig. 6, the model that considers all the different data sources outperforms the statistical model. Starting out more or less similarly, the model benefits from significant improvements as the season continues on, achieving an average score of 63 points per gameweek while the baseline (statistical) model only scored 52 points. Our model achieved 2314 points during testing, placing it within the top 30,000 players out of over 6 million players this season (top 0.5%). The baseline model reached 1994 points ranking it just below the 800,000 mark (top 13%).

Both models perform weakest when picking goalkeepers, with the statistical model performing slightly better having an overall precision of 82% against the multi-stream models 79%. This is likely because goalkeepers tend to be very consistent and heavily reliant on the team's overall form. This is shown in Fig. 5 which highlights the most important variables during goalkeeper predictions. On the contrary, when looking at forward predictions, both models perform exceptionally well, with the baseline predictor achieving a precision of 88% and the multi-stream predictor a precision of 92% when tested on individual gameweeks.

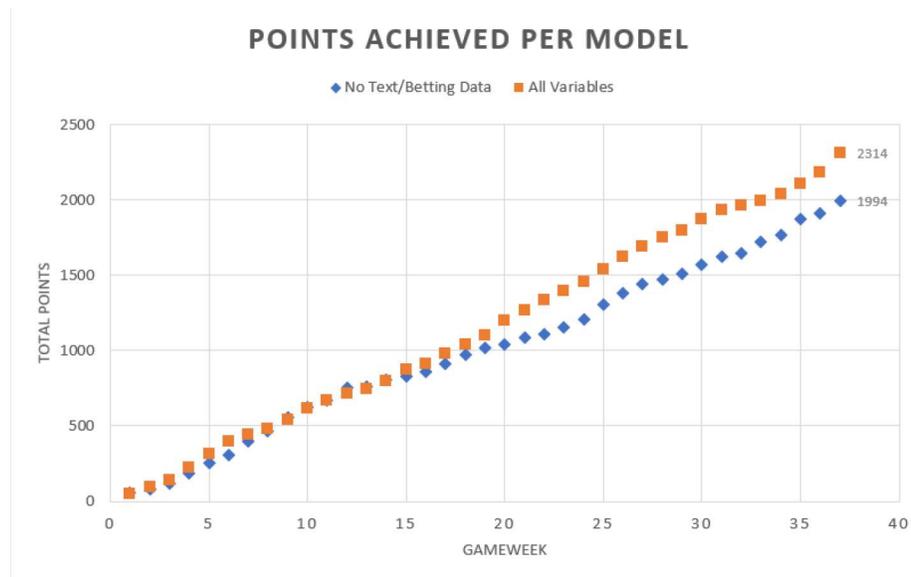

**Fig. 6.** FPL 2018/19 season overall points

We found that standard NLP libraries could not properly identify football-related sentiment in tweets. A corpus built using all the previous seasons FPL related tweets could have been one possibility to solve this issue, allowing us to incorporate tweets into our different models. Overall, we have observed significant improvements in this approach when compared to the baseline models, especially further into the season when more data is available. By properly understanding the strengths and weaknesses of each of the different data sources, the recommender system can be fine tuned to surpass the baseline models in all the positions. We now understand that statistics are much more important for goalkeeper selection than article recommendations, so a higher weighting should be applied in this scenario. In this research we show that Multi-stream data analytics is a viable concept and can be expanded to work not only for fantasy sports predictions but also for all other areas in which both statistics and human feedback are important factors; such as for example in stock markets or election voting predictions.

## 5 Conclusion

The primary research aim of this work was to investigate whether combining data from different data sources would help in increasing the accuracy and overall performance of predictions. Ultimately, we found that while the accuracy of predictions suffers slightly (from 91% to 89%), we found that the average points achieved per gameweek increases significantly from 54 to 64 points per gameweek. This is a significant points-increase, resulting in a rise in FPL leader board rankings from rank 800,000 (top 13%) to rank 30,000 (top 0.5%) out of the 6.5million total players. By properly understanding the strengths and weaknesses of each data-source, it becomes easy to solve such issues by simply applying more weight importance to statistical parameters when considering goalkeepers.

In this paper we aimed to combine statistical metrics with betting odds, news-articles & blogs, and Twitter data to improve FPL recommender services. We found that the inclusion of both betting data and news-articles & blogs individually, significantly improves on the power of predictions made by statistical predictors. However, by including both sets of data simultaneously, the predictions become considerably more powerful than their individual counterparts. This indicates that the incorporation of any relevant additional data would have positive effects on the predictive power of such recommender systems. While we also investigated adding social media data through Twitter, we faced many challenges leading to inaccurate predictions affecting the overall predictive power for the worse (Fig. 4. By creating a custom corpus based on the previous years of FPL data, FPL-specific social media could have a significantly beneficial impact on our model. This is because of the vast amount of discussions and independent opinions that are spread through such sites. Another alternative could have also been the FPL-specific Reddit discussion board [8].

Naturally, while this work was focused on fantasy football predictions, the methods described could be used in any domain that relies on statistics but also on human feedback; for example stock-markets or even political campaigns - both of which bring about lots of social-media data.

---
[8] https://www.reddit.com/r/FantasyPL/